\documentclass[runningheads]{llncs}
\usepackage[font=small,labelfont=bf]{caption}
\usepackage{multirow}  
\usepackage{etoc}
\usepackage{minitoc}

\usepackage{graphicx}

\usepackage{tikz}
\usepackage{comment}
\usepackage{amsmath,amssymb} 
\usepackage{cite}
\usepackage{color}
\usepackage{xspace}
\usepackage[font=small,labelfont=bf]{caption}

\usepackage[accsupp]{axessibility}  

\usepackage{algorithm}
\usepackage{algorithmicx}
\usepackage{algpseudocode}
\usepackage{amsmath}

\usepackage{xcolor}
\PassOptionsToPackage{hyphens}{url}\usepackage[pagebackref=true,breaklinks=true,letterpaper=true,colorlinks,
  citecolor=citecolor,bookmarks=false]{hyperref}
\definecolor{citecolor}{RGB}{34,139,34}

\def\themodel{PointTree\xspace}

\begin{document}
\pagestyle{headings}
\mainmatter
\def\ECCVSubNumber{}  

\title{\themodel: Transformation-Robust Point Cloud Encoder with Relaxed K-D Trees} 

\titlerunning{Transformation-Robust Point Cloud Encoder with Relaxed K-D Trees}
\author{Jun-Kun Chen\index{Chen, Junkun} \and Yu-Xiong Wang} %
\authorrunning{J.-K. Chen \and Y.-X. Wang}
\institute{University of Illinois at Urbana-Champaign\\
\email{\{junkun3, yxw\}@illinois.edu}}
\maketitle
\begin{abstract}
Being able to learn an effective semantic representation directly on raw point clouds has become a central topic in 3D understanding. Despite rapid progress, state-of-the-art encoders are restrictive to canonicalized point clouds, and have weaker than necessary performance when encountering geometric transformation distortions. To overcome this challenge, we propose {\em \themodel}, a general-purpose point cloud encoder that is {\em robust to transformations} based on {\em relaxed} K-D trees. Key to our approach is the design of the division rule in K-D trees by using principal component analysis (PCA). We use the structure of the relaxed K-D tree as our computational graph, and model the features as border descriptors which are merged with pointwise-maximum operation. In addition to this novel architecture design, we further improve the robustness by introducing {\em pre-alignment} -- a simple yet effective PCA-based normalization scheme. Our \themodel encoder combined with pre-alignment consistently outperforms state-of-the-art methods by large margins, for applications from object classification to semantic segmentation on various transformed versions of the widely-benchmarked datasets. Code and pre-trained models are available at \href{https://github.com/immortalCO/PointTree}{\texttt{https://github.com/immortalCO/PointTree}}.
\end{abstract}

\begin{figure}[ht]
\begin{center}
\centerline{\includegraphics[width=1.0\columnwidth]{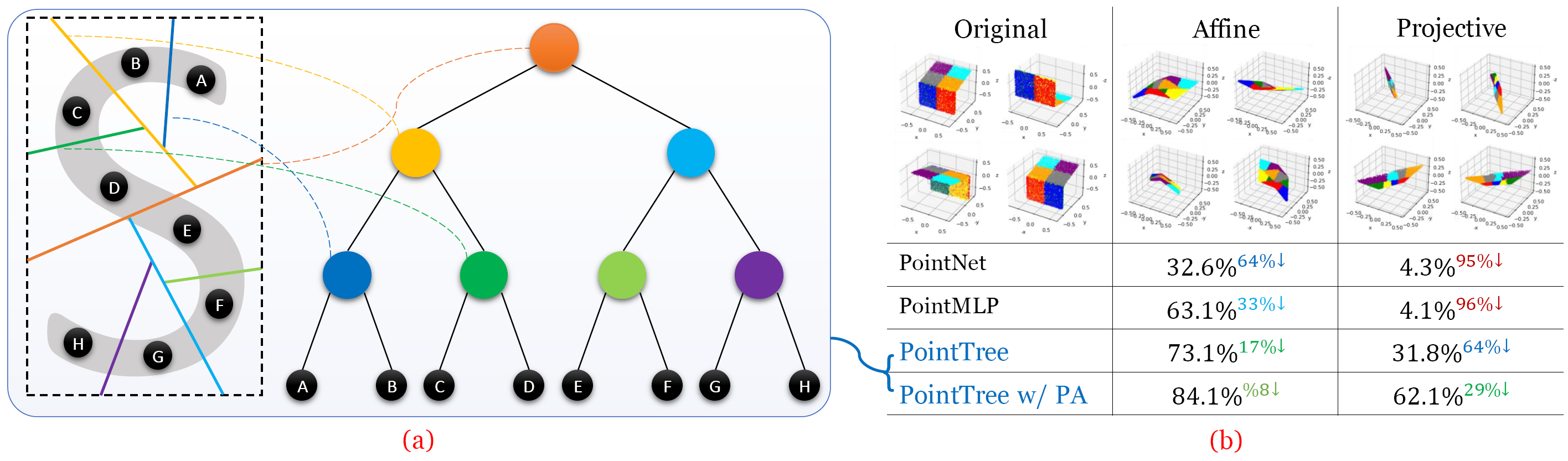}}
\caption{Our \textbf{robust} \themodel vs. \textbf{non-robust} existing point cloud encoders under geometric transformations. \textbf{\textcolor{red}{(a)}} \themodel is based on {\em relaxed} K-D trees, and its robustness is mainly achieved by designing the division plane of each node to be robust against transformations: An example of a relaxed K-D tree operated on an 8-point S-shaped point cloud. Each node is related to the division line with the same color, and each point is related to a leaf node. \textbf{\textcolor{red}{(b)}} Affine and projective transformations can highly deform the point clouds, dramatically decreasing the object classification accuracy of existing point cloud encoders like the widely-used PointNet~\cite{pointnet} and more recent PointMLP~\cite{pointmlp} (`$\downarrow$' indicates the relative performance drop compared with their accuracy on the canonicalized point clouds which is 90.6\% and 94.5\%, respectively). We illustrate an example point cloud in ModelNet40~\cite{modelnet}. For the original (canonicalized), affine, and projective versions, we show 4 different point of views of the same point cloud, and use different colors to represent different octants. \themodel significantly outperforms its counterparts in challenging transformation scenarios. With the additionally proposed pre-alignment (`PA'), \themodel further improves the accuracy}
\label{fig:teaser}
\end{center}

\end{figure}

\section{Introduction}
\label{sec:intro}
3D sensing technology has advanced rapidly over the past few years, playing a significant role in many applications such as augmented reality, autonomous driving, and geographic information systems~\cite{geosys,autodrive,augreal}. As one of the most commonly-used output formats of 3D sensors, 3D point clouds flexibly describe the surface information of the sensed objects or scenes with collection of points. Therefore, being able to learn an effective semantic representation directly on raw point clouds, which is useful for high-level tasks such as object recognition, has become a central topic in 3D understanding, with various powerful deep learning architecture based encoders emerging like PointNet~\cite{pointnet} and PointMLP~\cite{pointmlp}. In real-world applications, a desired encoder is supposed to cope with a wide range of {\em geometric transformations} -- point clouds of the same object category may undergo different similarity/affine/projective transformations, leading to large intra-class variations. This paper demonstrates that existing state-of-the-art point cloud encoders have weaker than necessary performance when encountering geometric transformation distortions, and proposes {\em \themodel}, a new type of encoder based on {\em relaxed} K-D trees that is {\em robust to transformations} and thus offers substantial improvements in performance.

Such investigation on transformation robustness for point cloud encoders is under-explored in existing work, partially because most of the commonly-used benchmark datasets, like ModelNet~\cite{modelnet} and ShapeNet~\cite{shapenet}, make a simplifying assumption -- the point clouds are in precise shapes and aligned in a canonical coordinate system. For example, a point cloud of a table in such a dataset always has its tabletop parallel to the xOy plane, and all its legs vertical to the top. However, such an assumption is restrictive and hardly satisfied {\em in the wild} -- e.g., an object scanned by a 3D sensor is often not aligned with the sensor, leading to unaligned point clouds. Due to noise, perturbation, viewpoint change, miscalibration, or precision limitation of the sensor itself~\cite{miscal1,miscal2,miscal3}, the scanned point cloud may have a different shape from the object, and the shape might be deformed by a 3D affine or projective transformation. On the other hand, most of existing encoders developed on these benchmarks highly rely on this assumption and are thus sensitive to input point cloud deformations, leading to dramatically degenerated recognition performance as shown in Figure~\ref{fig:teaser}\textcolor{red}{-b}. Notably, while effective for simple input corruptions, the common strategy (e.g., the T-Net~\cite{pointnet} used in PointNet) that adopts a transformer network to explicitly predict a transformation for canonicalization of input data cannot deal with more general deformations here.

To overcome this challenge, we propose \themodel, a transformation-robust, general-purpose point cloud encoder architecture. Key to our approach is the use of {\em relaxed} K-D trees~\cite{relaxedkdt}. While there have been some existing approaches~\cite{kdnetwork,3dcontextnet} that utilize K-D trees, they are based on the {\em conventional} K-D tree that only divides the point set along an axis at each node, which implicitly uses the aligned assumption and thus still performs poorly on transformed point clouds. By contrast, our use of the relaxed K-D tree removes the restriction of division rules, allowing more flexible designs. Particularly, in \themodel, we design the division rule by using principal component analysis (PCA) as illustrated in Figure~\ref{fig:teaser}\textcolor{red}{-a}. By doing so, theoretically, we show that our division of the point set and construction of the whole K-D tree are {\em invariant to similarity transformations}; and empirically, we observe that our \themodel exhibits strong {\em robustness against more complicated affine and projective transformations} (Figure~\ref{fig:teaser}\textcolor{red}{-b}). 

In addition to the proposed division rule, the robustness of our approach further stems from other properties of K-D trees and additional design strategies. As a tree structure containing multiple layers, \themodel natively divides a point cloud into components at bottom layers. This facilitates the recognition of multi-component objects (e.g., an airplane), as \themodel may still capture useful local features from lower layers even if the whole point cloud undergoes severe deformation. Also, the similarity transformation-invariant division induced by relaxed K-D trees prevents cutting two components with the same shape (e.g., two engines of the airplane) in different directions, so that the symmetricity can still be leveraged. Moreover, following PointNet~\cite{pointnet} and PointNet++~\cite{pointnet++}, we model the features in \themodel as border descriptors which can be merged with pointwise-maximum operation. We use the structure of the relaxed K-D tree as our computational graph, which contains a native locality clustering and down-sampling scheme. Finally, not only from this novel architecture design, but we also improve the robustness by introducing a simple PCA-based normalization scheme (called ``pre-alignment'') on input point clouds to \themodel. This is shown as a general normalization scheme that consistently and effectively improves the performance of other encoders under transformations as well.

{\bf Our contributions} are three-folds. (1) We propose \themodel, a general-purpose point cloud encoder architecture based on relaxed K-D trees, which is robust against geometric (affine and projective) transformations. (2) We introduce pre-alignment, a simple yet general PCA-based normalization scheme, which can consistently improve the performance of a variety of point cloud encoders under geometric transformations. (3) We show that our \themodel encoder combined with pre-alignment consistently outperforms state-of-the-art methods by large margins, on various transformed versions of ModelNet40~\cite{modelnet}, ShapeNetPart~\cite{shapenet}, and S3DIS~\cite{s3dis} benchmarks.

\section{Related Work}
\label{sec:related}

\noindent\textbf{Deep Learning on Point Clouds.} There are mainly four directions to build a deep learning model to process and analyze point clouds~\cite{survey}: (i) multi-layer perceptron (MLP) methods~\cite{pointnet,pointnet++,pointmlp} that use pointwise MLPs along with some multi-stage locality clustering and down-sampling; (ii) convolution methods~\cite{densepoint,pointcnn} that perform convolutions on voxels, grids, or directly in continuous 3D space; (iii) graph methods~\cite{dgcnn,pointgcn} that construct graphs with points as vertices and with neighborhood relations as edges, and apply graph models; and (iv) data structure methods~\cite{kdnetwork,3dcontextnet,voxelcontextnet,octnet} that use a hierarchical data structure like an OCTree or a K-D tree as the computational graph. Our \themodel belongs to a data structure method, since it uses a relaxed K-D tree as the computational graph. Intuitively, pure MLP methods (without locality clustering) are not robust against transformations, since they only rely on coordinate values; by exploiting locality, locality clustering and neighborhood graph may improve the robustness.

\noindent\textbf{Point Cloud Encoders Based on K-D Trees.} To the best of our knowledge, mainly four methods in the literature use K-D trees to build point cloud encoders: KD-Net~\cite{kdnetwork}, 3DContextNet~\cite{3dcontextnet}, PD-Net~\cite{pdnet}, and MRT-Net~\cite{mrtnet}, while more approaches exist based on OCTrees~\cite{octnet,voxelcontextnet,Xiang2019ANO,psinet}. KD-Net is a simple version of the K-D tree-based point cloud encoder -- it uses MLP to merge the information of two children nodes for each node. As an advanced version, 3DContextNet uses the border descriptor features from PointNet~\cite{pointnet} which can be merged by pointwise-maximum; it further proposes multi-stage training to exploit local and global context. PD-Net~\cite{pdnet} is a variant of KD-Net that replaces the vanilla K-D tree with a PCA-based K-D tree, but its feature aggregation still highly relies on coordinates. MRT-Net~\cite{mrtnet} uses the K-D tree only for preprocessing and uses convolutional layers for further modeling.

\noindent\textbf{Relation Between Our \themodel and Previous Models.} Instead of using conventional K-D trees, \themodel leverages relaxed K-D trees with a proposed division plane selection method, making it different from all existing K-D tree-based models. Similar to PointNet~\cite{pointnet} and 3DContextNet~\cite{3dcontextnet}, \themodel models the features as border descriptors. While the model design of \themodel is inspired by and similar to that of 3DContextNet in the ``feature learning stage,'' \themodel is simpler {\em without} relying on multi-stage training and local and global cues. Furthermore, \themodel has an extra alignment network as in PointNet.

\noindent\textbf{Investigation on Transformation Robustness.} 
While there has been interest in addressing robustness against geometric transformations of input point clouds, existing work mainly focuses on specific transformations like rotation and similarity. IT-Net~\cite{itnet} proposes a learnable normalization component (or ``alignment network'') which learns to recover the original point cloud. Other work~\cite{reqnn,lgrnet,rotinv,closerlook} proposes SO(3) (similarity transformation) robust, invariant, or equivariant architectures that maintain stable results when training on rotated point clouds. Shear transformation is also studied~\cite{robcuststudy}, which is a special type of affine transformation with deformation only performed on two of the three axes. None of the existing methods are able to cope with robustness against general affine or projective transformations as our work.

\label{sec:method}
\noindent\textbf{Problem Setting.} We design a deep learning model as a general-purpose point cloud encoder. A \textbf{point cloud} is an unordered set of points $P=\{p_i\}_{i=1}^n$. Each point $p_i$ is a 3-D vector $(x_i,y_i,z_i)$ representing the three coordinates. Our model directly takes $P$ as input, and outputs a set of vectors $O$. For a downstream task, another model takes $O$ as input, and outputs task-specific predictions.
\subsection{Point Cloud Encoder Based on Relaxed K-D Trees}
\label{sec:encoder}

\begin{figure}[ht]

\begin{center}
\centerline{\includegraphics[width=0.9\columnwidth]{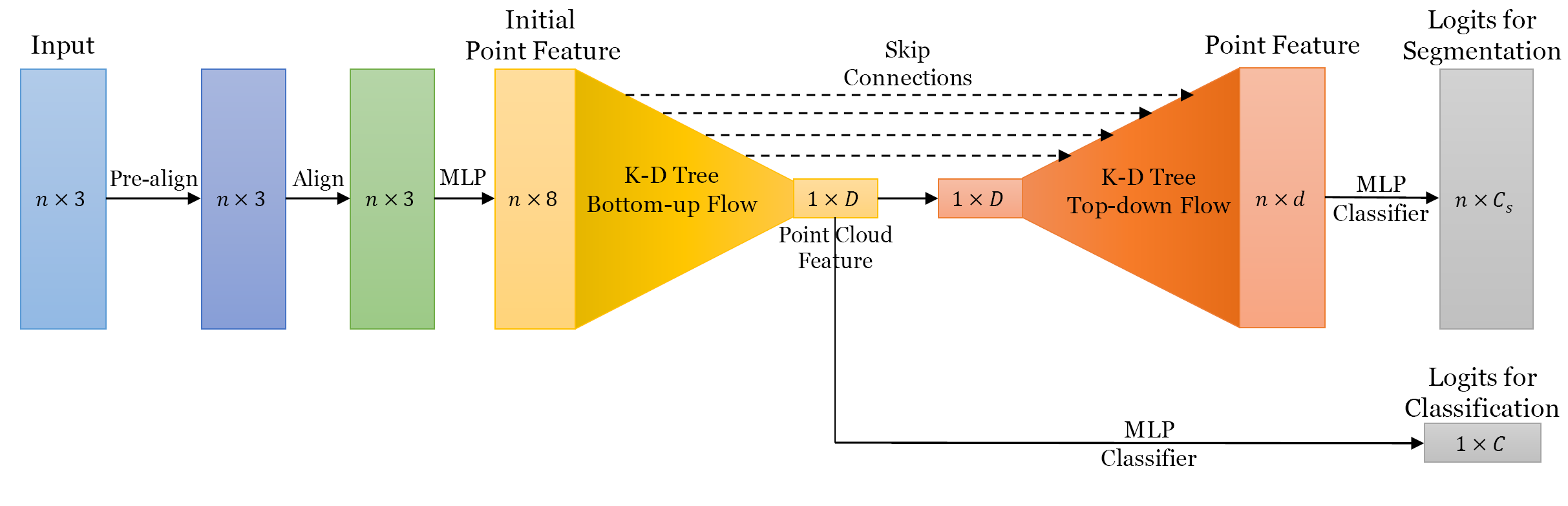}}
\caption{Architecture of \themodel. The input point cloud passes through a pre-alignment process, followed by an alignment network, and is then fed to our \themodel model. A K-D tree bottom-up flow is applied on such input, obtaining the point cloud feature which can be used for classification. For segmentation tasks, another K-D tree top-down flow is applied on the output of the bottom-up flow with some skip connections, obtaining point features which can be used for segmentation}
\label{fig:model}
\end{center}
\end{figure}

\noindent\textbf{K-D Trees and Relaxed K-D Trees.} Our proposed point cloud encoder {\em \themodel} is based on relaxed K-D trees, as illustrated in Figure~\ref{fig:model}. K-D trees are a classical data structure designed to solve $K$-dimensional range counting problems. It is a special decision tree built on $n$ $K$-dimensional input points, in two specific ways: (i) each node has an axis-parallel criterion; and (ii) such a criterion strictly divides the input points that go through this node to two equal-size parts. Each leaf node is related to exactly one input point. 

A K-D tree of depth $d$ is a full binary tree with $2^d$ points. The root is at layer $0$; the leaves are at layer $d$. Each non-leaf node $o$ has two unordered children nodes $o_l$ and $o_r$, while each leaf node has a corresponding point $p(o)$. Each non-root node has a unique parent node $\textbf{par}(o)$. At each non-leaf node, a linear criterion $W_o p + b_o \le 0$ divides the point set into two subsets, which is recursively processed at the left and right nodes. Hence, the subtree of each node contains the points in a continuous 3-D space, leading to a native locality clustering.

Existing K-D tree-based methods~\cite{kdnetwork,3dcontextnet} use the conventional definition of K-D trees that only divides along one axis, i.e., $W_o\in \{(0,0,1),(0,1,0),(1,0,0)\}$. However, such methods are not exploiting the expressiveness of K-D trees. By contrast, we adopt \textbf{relaxed K-D trees}~\cite{relaxedkdt} -- a generalization of K-D trees by removing the restriction of division plane, so that the point set can be divided with any criterion. Here, to improve the transformation robustness, we consider an arbitrary linear criterion, i.e., $W_o$ can be an arbitrary vector.

\begin{figure}[t]
\begin{center}
\centerline{\includegraphics[width=0.7\columnwidth]{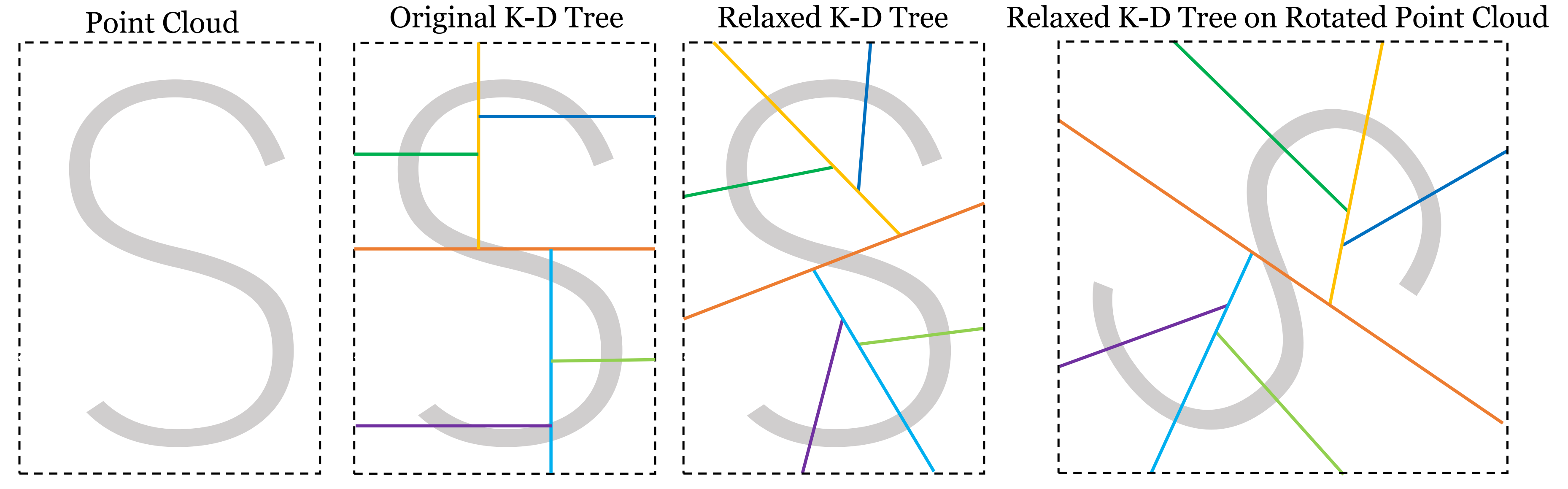}}
\caption{Relaxed K-D tree's \textbf{division and similarity equivariance}. For the S-shape point cloud, the original K-D tree divides it into a large number of {\em imbalanced} segments (one of the parts even contains two disconnected segments of S), while the relaxed K-D tree can cut it into a small number of {\em balanced} segments. Also, when rotating the S-shape with an angle, each division plane and pieces of the point cloud divided by the relaxed K-D tree rotate with the same angle and each divided segment keeps unchanged. This shows that relaxed K-D trees are equivariant to similarity transformations}
\label{fig:general-kdt}
\end{center}
\end{figure}

Concretely, \themodel uses the first principle component (obtained by a PCA algorithm) as the division plane, and chooses the medium value as the division boundary. As PCA is similarity-transformation equivariant, this K-D tree construction has the invariance against similarity transformations, as shown in Figure~\ref{fig:general-kdt}. Importantly, this property also brings in strong robustness against affine and projective transformations, as empirically validated in Section~\ref{sec:exp}. In addition, if the point cloud contains multiple similar components due to repeating or symmetricity, e.g., airplane engines, table legs, and desk drawers, our similarity-transformation invariant K-D tree construction can divide these components in the same way, further improving the model robustness and facilitating feature extraction from these components.

\noindent\textbf{Bottom-Up Information Flow.} In a traditional K-D tree algorithm, there is a scheme to upload information in a bottom-up way, called \textbf{bottom-up information flow}. In such information flow, each node $o$ takes some information as a vector $\mathrm{info(o)}$. The information at each leaf node is derived from the corresponding point, and the information at each non-leaf node is the aggregated information of its children nodes, as the formula below:
\begin{equation}
\end{equation}

The uploading process of the information flow is a natural down-sampling scheme. In the layer-by-layer bottom-up uploading process, the number of nodes in layer $i$ (which is $2^i$) is decreasing, and the size of the subtree of each node (which contains information of $2^{d-i}$ points) is increasing. Thus, we can regard each layer as a stage of down-sampling with down-sampled points. Such down-sampling process starts with the information of original points $P$ and ends with a single point $\mathrm{info}(R)$, where $R$ is the root of the K-D tree.

\begin{figure}[t]
\begin{center}
\centerline{\includegraphics[width=0.95\columnwidth]{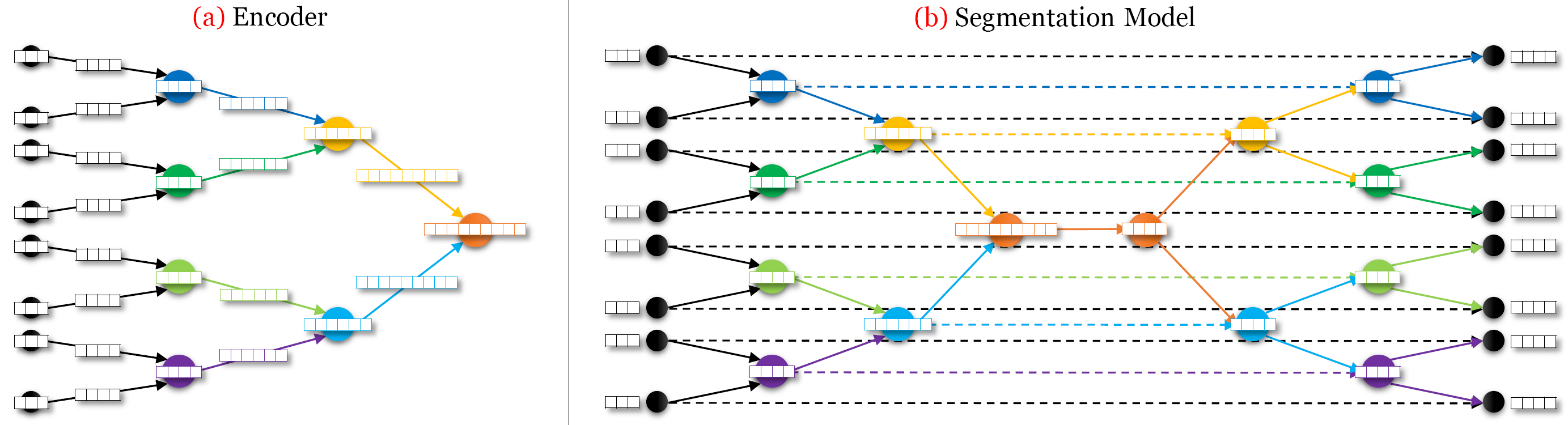}}
\caption{\textbf{\textcolor{red}{(a)}} \textbf{Our \themodel encoder}. The encoder has a multi-stage down-sampling and dimension increasing scheme, where each layer in the K-D tree is a stage. The points are down-sampled through layers via a bottom-up information flow. The grids on the nodes and edges show the dimension of features. For each node, the features from its two children nodes are first fed to a dimension-increasing MLP, and then merged with pointwise-maximum operation. \textbf{\textcolor{red}{(b)}} \textbf{The segmentation model}. The two symmetric K-D trees represent two stages: a bottom-up information flow (\textbf{left, the same as \textcolor{red}{(a)} with simplified notation}) and a top-down information flow (\textbf{right}), on the same K-D tree. The information is first uploaded from leaves to root to obtain global features, and then downloaded from root to leaves, obtaining the relationship between each leaf/point and the whole point cloud which is used for segmentation classification. For the top-down information flow (\textbf{right}), each node has a carried feature that represents the information of the relationship between its subtree and the whole point cloud. Such feature is obtained by merging its parent nodes' carried information and with a skip connection (dotted lines) from the same node in the bottom-up feature}
\label{fig:encoder_segment}
\end{center}
\end{figure}

\noindent\textbf{Our Encoder.} We design the encoder of \themodel (see Figure \ref{fig:encoder_segment}\textcolor{red}{-a}) by leveraging the idea of PointNet~\cite{pointnet} and its improved version PointNet++~\cite{pointnet++}, but in the information flow of a relaxed K-D tree. Specifically, we model the information as the border descriptor or ``global shape descriptor'' defined in PointNet~\cite{pointnet}, which can be aggregated with pointwise-maximum operation (a.k.a. maximum pooling). For each node $o$, the information or feature $\mathrm{info}(o)$ is the coarse shape descriptor of the points in its subtree.

We compute the information of each node layer-by-layer in a bottom-up order. For leaf nodes, the information is obtained by applying an MLP on the coordinates. For non-leaf nodes, the information is obtained by processing their left and right children nodes with a dimension-increasing linear transformation (defined on each layer), followed by aggregation with a pointwise-maximum operation, as the formula below:
\begin{equation}
\mathrm{info}(o) = \left\{ \begin{array}{lr} \textbf{MLP}(p(o)), &o\in L_d, \\ \textbf{pointwise-max}(W_i\mathrm{info}(o_l), W_i\mathrm{info}(o_r)), &\text{otherwise}.\end{array} \right. \label{equ:info-agg-pwmax}
\end{equation}

The encoder of \themodel can be regarded as a K-D tree-guided version of PointNet++, where at each stage of down-sampling, the points are clustered under some rule, and each cluster is down-sampled to one point with higher dimension. In \themodel, instead of some ad hoc clustering strategies used by PointNet++, K-D trees provide a native and principled way to cluster according to the spatial and neighborhood information, making it more powerful, general, and reliable. The output of the \themodel encoder is defined as $O=\{\mathrm{info}(o)\mid o\}$. 

\subsection{Robustness Against Transformations}
\label{sec:robustsub}
The robustness of \themodel against transformations mainly stems from our design of the division rule. Here, we discuss the robustness in more detail, introduce additional strategies that further improve the robustness, and propose a metric that quantifies the transformation intensity.

\noindent\textbf{Similarity Transformation.} \themodel uses relaxed K-D trees as its base tree, which holds the following lemma (the proof is in the supplementary material):

\noindent \textbf{Lemma.} If the rule to choose the division plane at each node is equivariant to similarity transformation $\sigma$, or more formally,
\begin{equation}
\textbf{choose-division-plane}(\sigma(P)) = \sigma(\textbf{choose-division-plane}(P)),
\end{equation}
where $\textbf{choose-division-plane}(P)$ is the procedure to choose the division plane on point set $P$, then the construction of the relaxed K-D tree is invariant to such a similarity transformation. 

\themodel uses PCA to implement $\textbf{choose-division-plane}(P)$, which is equivariant under any similarity transformation. As a result, our model is natively invariant to similarity transformations.

\noindent\textbf{Affine (and Projective) Transformations.} Interestingly, as empirically validated in Section~\ref{sec:exp}, this invariance against similarity transformations enables \themodel also highly robust against affine and even more complicated projective transformations. To further improve the robustness to affine transformations, we introduce two additional strategies: pre-alignment and alignment network. Correspondingly, an input point cloud is fed forward the pre-alignment process and then the alignment network, before passing to our \themodel encoder (Figure~\ref{fig:model}).

\noindent\textbf{Pre-alignment: A Normalization for Affine Transformations.} We design a PCA-based pre-alignment scheme as ``normalization" of affine transformed point clouds. For a centered point cloud $P\in \mathbb{R}^{n\times 3}$, applying PCA obtains $P = U\textbf{diag}(\Sigma) V^T$, where $U$ is a $3\times 3$ matrix, $\Sigma$ is a length-$3$ vector which can be regarded as scalings of each axis, and $V$ is an orthorgonal $3\times 3$ matrix which can be regarded as a rotation. When we apply an affine transformation to a point cloud, the scaling and rotation can be arbitrary, and thus the normalization should not take these two pieces of information. So we disregard them in $\Sigma$ and $V$, and take $U$ as the normalized or pre-aligned point cloud (equivalent to normalizing $P$ by applying another affine transformation $V\textbf{diag}(\Sigma^{-1})$). {\em Notably}, this pre-alignment scheme does not rely on the properties of K-D trees. As shown in Section~\ref{sec:exp}, it is a general approach and can be used for a variety of existing point cloud encoders to improve their robustness. In our implementation, we found that applying pre-alignment {\em iteratively} further improves the performance, especially for part segmentation. Also, we have proven empirically that such a pre-alignment method is invariant to affine transformations. See the supplementary material for more details.

\noindent\textbf{Alignment Network: A Learnable Component for Alignment.}
The pre-alignment is simply unlearnable. We also propose a {\em learnable} alignment network, inspired by PointNet~\cite{pointnet}. This network takes a feature vector of a point cloud as input, feeds it into an MLP, and outputs a length-$9$ vector, which is reshaped to a $3\times 3$ affine matrix to align the points. The encoder to generate the feature vector and the MLP are the learnable components of the alignment network.

The alignment network supports any encoder that outputs a feature vector, like a PointNet, another \themodel encoder, etc. Our default model uses the same architecture as ``T-Net'' in PointNet~\cite{pointnet}. Note that such an alignment network is {\em not} designed for restoring the original point cloud (as a registration task). Instead, we only expect that it can learn to convert the input point cloud into an easier form for the following \themodel encoder. In Section~\ref{sec:allvars}, we also consider other variants of the alignment network that adopt different architectures.

\noindent\textbf{Transformation Intensity Metric: Expected Angle Difference (EAD).} To evaluate the intensity of a transformation, we propose a metric called expected angle difference (EAD). It is defined on two point clouds $P$ and $P'$ as follows: if we uniformly sample three different point indices $a,b,c\in [|P|]$, then the EAD is defined as the expected difference of $\angle P_bP_aP_c$ and $\angle P'_bP'_aP'_c$. Or, 
\begin{equation}
\textbf{EAD}(P,P') = \mathbb{E}_{a,b,c\in [|P|]} \left[ \textbf{angle-diff}(\angle P_bP_aP_c, \angle P'_bP'_aP'_c) \right].
\end{equation}
EAD is a metric for measuring the deformation of the transformed point cloud. By definition, similarity transformations hold $\textbf{EAD}(P, P')=0$, representing the minimum deformation -- no deformation. Also, for the affine transformation which we randomly generated, the EAD is approximately $\frac{\pi}{8}$ (supplementary material). Meanwhile, the experiment shows that our pre-alignment scheme yields 
\begin{equation}
    \textbf{EAD}(\textbf{pre-align}(\textbf{affine}_1(P)), \textbf{pre-align}(\textbf{affine}_2(P))) < 10^{-4},
\end{equation}
for any two affine transformations $\textbf{affine}_1$ and $\textbf{affine}_2$. This indicates that our pre-alignment is effective, which can normalize different affine transformations on a same point cloud to similar point clouds. More analysis about the EAD of transformations and pre-alignment methods are in the supplementary material.

\subsection{Downstream Components}

\noindent\textbf{Classification: Point Cloud Features.} In a classification task, each point cloud belongs to one class. Given a point cloud, the model should predict its class within all class candidates. For \themodel, the root node's information accounts for the whole point cloud, and we treat it as a global feature. We then build an MLP classifier that takes the root information $\mathrm{info}(R)$ as input. The output will be the log-likelihood scores for $C$ candidate classes (Figure~\ref{fig:model}).

\noindent\textbf{General Segmentation: Point Features with Top-Down Information Flow.}
In a general segmentation (e.g., part or semantic segmentation) task, for a given point cloud, each point belongs to one of $C_S$ candidate classes. The model should classify all points in the given point cloud. We design a segmentation decoder following KD-Net~\cite{kdnetwork}, as shown in Figure~\ref{fig:encoder_segment}\textcolor{red}{-b}. The decoder is a K-D tree symmetric to the encoder. It follows a top-down flow, as opposite to the bottom-up flow in the encoder. Every node in the decoder has a feature called ``carried information,'' representing the global-local relationship between inside and outside its subtree. Therefore, we can model the ``role'' of the subtree in the global shape. And when the node is leaf, it is exactly the ``role'' of the corresponding point in the global shape, which can be viewed as point feature.

Each node takes two inputs: the carried information from its ancestor, and the skip connection from the symmetric node in the encoder. The node merges these two inputs with one MLP, obtaining the carried information of itself. The top-down flow ends at leaf nodes and outputs the carried information of leaves. Such information is the feature of each point and is used for segmentation.

\subsection{\themodel Variants}
\label{sec:allvars}

We introduce three variants of \themodel with different design of alignment networks and encoders. Note that, as mentioned in Section~\ref{sec:robustsub}, the alignment network supports any encoder that outputs a point cloud feature. (1) \textbf{Default encoder (`Def')} uses T-Net in PointNet~\cite{pointnet} as the alignment network. (2) \textbf{Encoder with K-D tree alignment (`KA')} introduces the default encoder as a stronger alignment network.  (3) \textbf{ResNet-style encoder (`RNS')} is a ResNet-Style variant of \themodel with increased model capacity, by stacking more layers in a ResNet's style~\cite{resnet}. By connecting a default encoder and a 
general segmentation component, we can convert the $N\times 3$ input features to $N\times d$ intermediate features. We define such a connected structure as a ``ResNet block,'' and stack a block followed by a default encoder to build a ResNet-style encoder. The output of each block is linked with the output of the previous block through a skip connection as in ResNet. For this variant, we can also treat the last encoder as the main encoder, and all previous encoders as part of the alignment network (since they mostly affect the input at the last layer of the last encoder). The detailed architectures are shown and explained in the supplementary material.

\label{sec:exp}
\noindent\textbf{Transformations.} We evaluate our model on affine and projective transformed versions of existing datasets, including ModelNet40~\cite{modelnet}, ShapeNetPart~\cite{shapenet}, and S3DIS~\cite{s3dis}. For a dataset $D = \{P\}$ and a random distribution $T$ of transformations, we construct the $T$-transformed dataset as follows: for each $P\in D$, we sample a fixed number (``augment time'') of transformations $\{t\}$, and add all $t(P)$ in the dataset. Different point clouds will be applied to different transformations. We perform such process for $D_{\mathrm{train}}$, $D_{\mathrm{val}}$, and $D_{\mathrm{test}}$ separately with some specific ``augment time,'' obtaining a full transformed dataset.

\noindent\textbf{Baselines and \themodel Variants.} For baseline models, we run the experiments with their official code by injecting the transformations into their data loaders. We keep their original optimal hyper-parameters, and train the models until convergence. We focus on four types of baselines: (i) PointNet related models (PointNet~\cite{pointnet} and PointNet++~\cite{pointnet++}), (ii) K-D tree-based models (KD-Net~\cite{kdnetwork}, 3DContextNet~\cite{3dcontextnet}, and PD-Net~\cite{pdnet}), (iii) recent state-of-the-art models (DGCNN~\cite{dgcnn}, GBNet~\cite{gbnet}, GDANet~\cite{gdanet}, CurveNet~\cite{curvenet}, and PointMLP~\cite{pointmlp}), and (iv) SO(3) robust models (CloserLook~\cite{closerlook} and LGR-Net~\cite{lgrnet}). As some baselines do not release their code on segmentation tasks, we only evaluate them on the classification task. We evaluate all three \themodel variants (Section \ref{sec:allvars}).

\noindent\textbf{Classification on ModelNet40.} For the classification task, we evaluate our model on ModelNet40~\cite{modelnet}. ModelNet40 contains 9,843 point clouds for training and 2,468 point clouds for testing, and each of them belongs to one of 40 categories. We run the experiment on affine and projective transformations, and report the overall accuracy as our metric.

\begin{table}[t]
\centering
\caption{\themodel significantly outperforms state-of-the-art point cloud encoders under affine transformations for object classification (instance-level overall accuracy (\%)) on the affine transformed ModelNet40 dataset. With the proposed pre-alignment (`PA'), the performance of all methods consistently improves, and \themodel still achieves the best result. The results of baselines are obtained by running publicly released code on the transformed dataset. In addition, \themodel's accuracy has a lower standard deviation on different affine transformed datasets (supplementary material)}
\setlength{\tabcolsep}{4pt}
\scalebox{0.8}{\begin{tabular}{c|l|cc}
 \hline
 Type & Method & Affine w/ PA & Affine w/o PA \\
 \hline
 \multirow{2}{*}{PointNet related} & PointNet~\cite{pointnet}  & 51.1 & 32.6\\
 & PointNet++~\cite{pointnet++}  & 72.7 & 47.8\\
 \hline 
 \multirow{3}{*}{K-D Tree-based} & KD-Net~\cite{kdnetwork}  & 65.3 & 23.1 \\
 & 3DContextNet~\cite{3dcontextnet}  & 76.7 & 37.1 \\
  & PD-Net~\cite{pdnet}  & 62.0 & 25.7 \\
 \hline 
 \multirow{6}{*}{State-of-the-art} & DGCNN~\cite{dgcnn}  & 79.4 & 57.4\\
 & GBNet~\cite{gbnet}  & 69.4 & 18.7 \\
 & GDANet~\cite{gdanet}  & 72.8 & 15.6 \\
 & CurveNet~\cite{curvenet}  & 82.1 & 59.3 \\
 & PointMLP~\cite{pointmlp}  & 82.3 & 63.1 \\
 & IT-Net~\cite{itnet} + DGCNN~\cite{dgcnn} & 80.6 & 64.2\\
 \hline
  \multirow{2}{*}{SO(3) Invariant/Equivariant} & CloserLook\cite{closerlook} & 82.4 & 64.9 \\
  & LGR-Net\cite{lgrnet} & 80.1 & 62.7 \\
 \hline
 \multirow{1}{*}{Ours} &\themodel RNS  & \bf{84.1} & \bf{73.1}\\
 \hline
\end{tabular}}

\label{tab:exp-cls-affine}
\end{table}

\emph{Affine Transformations.} As shown in Table \ref{tab:exp-cls-affine}, our robust \themodel consistently outperforms all other point cloud models. Notably, \themodel significantly outperforms other models in the setting of affine without pre-alignment by more than 7\%$\sim$10\%. This clearly shows that \themodel has much higher robustness against affine transformations. Our accuracy is also 8\% and 36\% higher than 3DContextNet~\cite{3dcontextnet}, the previous best K-D tree-based model, in settings of affine with and without pre-alignment, respectively. This validates that the relaxed K-D tree is crucial to achieving the robustness that the original K-D tree is unable to. In addition, PD-Net~\cite{pdnet} uses a similar PCA-based relaxed K-D tree as ours, but its design of feature aggregation highly relies on coordinates (by using the normal vector of children nodes' division plane), which wastes and nullifies the affine robustness of the tree structure, yielding a 13.6\% worse accuracy than ours. Our method also consistently outperforms the SO(3) robust baselines, showing that the SO(3) robustness is not sufficient for coping with affine transformations. 

Finally, by comparing the settings of affine with and without pre-alignment, we observe an at least 10\% improvement in accuracy when applying pre-alignment in each baseline. This shows  that pre-alignment is a general and effective approach to normalizing affine transformed point clouds.

\begin{table}[t]
\centering
\caption{\themodel is robust even on the highly-challenging projective transformed ModelNet40 dataset, with and without pre-alignment (`PA'). It significantly outperforms all other models with a huge gap of 25\% at instance-level accuracy (\%)}
\setlength{\tabcolsep}{4pt}
\scalebox{0.85}{\begin{tabular}{l|ccc}
 \hline
 Method & Projective w/ PA & Projective w/o PA \\
 \hline
PointNet & 15.4 & 4.3 \\
\hline
DGCNN & 47.3 & 6.2 \\
PointMLP & 49.9 & 4.1\\
CurveNet & 37.6 & 5.6 \\
 \hline
 \themodel RNS & \bf{62.1} & \bf{31.8}\\
 \hline

\end{tabular}}

\label{tab:exp-cls-homo}
\end{table}

\emph{Projective Transformations.} Table \ref{tab:exp-cls-homo} shows the accuracy of our model and top baselines on the projective transformed ModelNet40 dataset. In this experiment, \themodel significantly outperforms all baselines by more than 25\%. For the most challenging setting, projective ModelNet40 without pre-alignment, \themodel still achieves a reasonable accuracy, while all other baselines can only obtain an accuracy that is a little higher than random guess.

\begin{table}[t]
\centering

\caption{Ablation study results show that both pre-alignment (`PA') and alignment network improve our accuracy on ModelNet. Among the three variants, RNS achieves the best performance, but all of them outperform baselines in Table~\ref{tab:exp-cls-affine}. Also, relaxed K-D tree is crucial for the transformation robustness in \themodel}
\setlength{\tabcolsep}{4pt}
\scalebox{0.85}{\begin{tabular}{lll|c}
 \hline
 Method & & & ModelNet40 Affine \\
 \hline
 \themodel RNS &w/ PA& & 84.1\\
 \themodel KA &w/ PA& & 83.4 \\
 \themodel Def &w/ PA& & 82.7 \\
 
 \hline
 \themodel RNS &w/o PA& & 73.1\\
 \themodel KA &w/o PA& & 71.7 \\
 \themodel Def &w/o PA& & 70.2 \\
 \hline
 \themodel Def &w/ PA &w/o Alignment Network & 82.4 \\
 \themodel Def &w/ PA& w/ Original K-D Tree & 68.8 \\

 \hline 
 \themodel RNS &w/ PA &w/ Concatenate-MLP & 79.3 \\
\hline
 
\end{tabular}}
\label{tab:exp-cls-abla}
\end{table}

\emph{Ablation Study.} From the ablation study results in Table~\ref{tab:exp-cls-abla}, we have the following observations. 
(1) By comparing the accuracy of the three \themodel variants in settings with and without pre-alignment, we observe that pre-alignment is helpful for our already-robust \themodel.
(2) The three variants -- Def, KA, and RNS -- can be also interpreted as different types of the ``alignment network'' component in \themodel, from simplest to the most sophisticated. By comparing their accuracy, along with the variant ``\themodel Def w/o Alignment Network,'' we can find that the improvement in alignment networks leads to higher accuracy.
(3) RNS is the most powerful variant, due to its multiple layers and intermediate features. {\em Notably}, even the other variants have lower accuracy than RNS, they still outperform all the baselines in Table \ref{tab:exp-cls-affine}.
(4) When we replace the relaxed K-D tree with the original K-D tree in \themodel, the performance experiences a dramatical drop by more than 10\%. This indicates that the relaxed K-D tree is crucial for the robustness against transformations. 
(5) When we replace the implementation of $\textbf{merge-info}$ from pointwise-maximum in Formula \eqref{equ:info-agg-pwmax} to concatenate-MLP (concatenate $o_l$ and $o_r$ and apply an MLP), the accuracy clearly drops, showing that pointwise-maximum is a critical design choice.

\begin{figure}[ht]
\begin{center}
\centerline{\includegraphics[width=0.9\columnwidth]{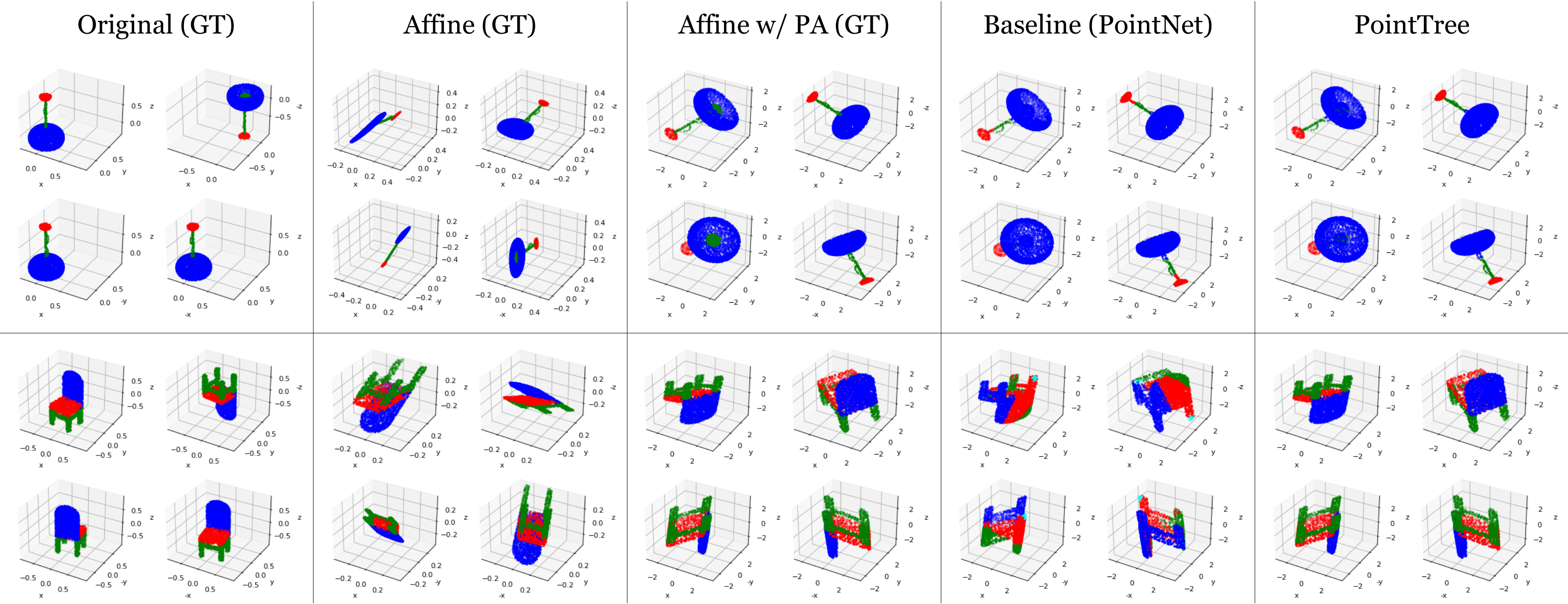}}
\caption{Visualization of two point clouds in ShapeNetPart, showing that pre-alignment can successfully normalize the highly-deformed point clouds under affine transformations into reasonable shapes, and that \themodel works almost perfectly in this setting. Each grid of the table contains 4 point of views of a same point cloud. The third column shows the pre-aligned point clouds of the second column. The point clouds in first 3 columns are colored according to ground truth (`GT') segmentation, while last 2 columns are colored according to the segmentation outputs of PointNet and \themodel}
\label{fig:exp-seg-visu}
\end{center}
\end{figure}

\noindent\textbf{Part Segmentation on ShapeNetPart.} We test \themodel for the point cloud part segmentation task on ShapeNetPart~\cite{shapenet}. It contains 16,881 point clouds in 16 classes. Each point belongs to one of 50 parts, where different classes have different sets of parts. We report the class-level mean intersection over union (mIoU) as the accuracy. Figure \ref{fig:exp-seg-visu} visualizes two point clouds with \themodel and PointNet~\cite{pointnet} as baseline. In both cases, the pre-alignment successfully normalizes the very flat affine transformed point cloud into a reasonable shape. For the lamp case, both PointNet and \themodel make a small mistake at the center of the lamp top (which is marked blue but should be green), while \themodel is more accurate at the bottom of the light pole. For the chair case, PointNet is performing badly, while our \themodel's accuracy is almost perfect.

\begin{table}[t]
\centering
\caption{\themodel significantly outperforms baselines for part segmentation on affine transformed ShapeNetPart with pre-alignment, increasing class-level mIoU (\%) by 7\%}

\setlength{\tabcolsep}{1pt}
\scalebox{0.73}{\begin{tabular}{l|cccccccccccccccc|c}
 \hline
 Method & {\tiny airplane} & {\tiny bag} & {\tiny cap} & {\tiny car} & {\tiny chair} & {\tiny earphone} & {\tiny guitar} & {\tiny knife} & {\tiny lamp} & {\tiny laptop} & {\tiny motorbike} & {\tiny mug} & {\tiny pistol} & {\tiny rocket} & {\tiny skateboard} & {\tiny table} &  mIoU \\
 \hline
 PointNet~\cite{pointnet} &69.7 &53.5 &57.8 &59.2 &84.1 &44.9 &62.9 &41.7 &61.9 &64.5 &31.8 &71.0 &50.3 &35.4 &46.1 &76.6 & 56.9 \\
  PointNet++~\cite{pointnet++}&66.5 &73.7 &58.6 &37.6 &72.0 &71.5 &85.9 &74.4 &72.5 &51.0 &29.1 &74.9 &54.1 &41.6 &57.1 &71.9 & 62.1 \\
 \hline 
 DGCNN~\cite{dgcnn} &\bf{89.0} &71.3 &\bf{91.2} &33.7 &11.7 &\bf{94.3} &84.6 &67.5 &72.6 &\bf{92.9} &10.3 &84.1 &\bf{83.7} &35.1 &62.9 &\bf{97.3}  & 67.6 \\
 GDANet~\cite{gdanet} &76.4 &74.3 &78.3 &59.1 &84.4 &72.7 &86.7 &75.9 &74.0 &70.0 &31.1 &89.5 &65.7 &53.8 &\bf{75.6} &79.5  & 71.7 \\
 CurveNet~\cite{curvenet} &75.8 &64.7 &79.3 &60.7 &86.9 &59.5 &86.2 &72.4 &74.8 &69.3 &24.6 &89.5 &63.5 &39.2 &60.1 &78.4  & 67.8 \\
 \hline
 \themodel RNS &83.6 &\bf{74.7} &82.5 &\bf{80.4} &\bf{90.6} &66.5 &\bf{92.1} &\bf{84.0} &\bf{82.0} &88.3 &\bf{56.0} &\bf{95.4} &78.0 &\bf{55.4} &68.9 &81.2 &\bf{78.7} \\
 \hline
\end{tabular}}

\label{tab:exp-seg-affine}

\end{table}

Table \ref{tab:exp-seg-affine} shows the segmentation results on affine transformed ShapeNetPart with pre-alignment. Our \themodel achieves a top mIoU over all baselines with an increase of more than 7\%, and achieves best IoUs for more than half of the classes. This shows that \themodel is a general-purpose encoder that can work in both classification and segmentation tasks, being significantly more robust than other models. The results on projective transformed ShapeNetPart in the supplementary material demonstrate similar observations.

\begin{table}[t]
\centering
\caption{\themodel consistently outperforms state of the art for large-scale semantic segmentation on S3DIS, with notable margins of 2.2\% and 6.7\% on affine and projective transformed datasets, respectively, in overall accuracy (\%) of Area5}
\setlength{\tabcolsep}{4pt}
\scalebox{0.85}{\begin{tabular}{l|cccc}
 \hline
 Method & Affine w/ PA & Affine w/o PA & Projective w/ PA & Projective w/o PA \\
 \hline
PointNet~\cite{pointnet} &64.9 &69.1 &49.3 &31.1 \\
DGCNN~\cite{dgcnn} & 80.0 & 70.8 &67.1 & 57.8 \\
 \hline
 \themodel Def & \bf{82.2} & \bf{74.9} & \bf{73.8} &\bf{61.2}\\
 \hline 
 
\end{tabular}}
\label{tab:exp-s3dis}
\end{table}

\noindent\textbf{Semantic Segmentation on S3DIS.} Table \ref{tab:exp-s3dis} shows the semantic segmentation results on affine and projective transformed S3DIS~\cite{s3dis} in settings with and without pre-alignment. Our \themodel achieves a top Area 5 overall accuracy over all baselines by large margins of at least 2.2\%. \themodel is thus not only robust in simple single-object point cloud tasks like ModelNet40 and ShapeNetPart, but is also robust in complicated multi-object point cloud tasks like S3DIS.

\section{Conclusion}
In this paper, we proposed \themodel, a general-purpose point cloud encoder that is highly robust against affine and projective transformations. The key insight of \themodel is the use of relaxed K-D trees with PCA-induced similarity transformation-invariant construction. We further introduced pre-alignment, an effective and model-agnostic normalization scheme. Empirical evaluation shows that \themodel  significantly outperforms state-of-the-art methods on various transformed datasets for classification and segmentation tasks. Notably, under affine transformations, the combination of \themodel with pre-alignment even achieves an accuracy that is close to the accuracy on the canonicalized point clouds. We hope our work could inspire more efforts on developing robust point cloud analysis models, and promote better exploitation of powerful K-D trees.

\noindent\small{\textbf{Acknowledgement.} This work was supported in part by NSF Grant 2106825, the Jump ARCHES endowment through the Health Care Engineering Systems Center, the New Frontiers Initiative, the National Center for Supercomputing Applications (NCSA) at the University of Illinois at Urbana-Champaign through the NCSA Fellows program, and the IBM-Illinois Discovery Accelerator Institute.}

\chapter*{Supplementary Material}
\appendix
\renewcommand{\thefigure}{\Alph{figure}}
\renewcommand{\thetable}{\Alph{table}}
\renewcommand{\thealgorithm}{\Alph{algorithm}}

This document contains additional descriptions (e.g., formal or detailed definition, theoretical proofs, implementation details, etc.) and extra experiments (e.g., segmentation task under projective transformation, overtime accuracy, stability test, etc.). 

\section{Formal Definitions of K-D Trees}

This section extends \textbf{Section 3.1 Point Cloud Encoder Based on Relaxed K-D Trees} in the main paper with formal and detailed definitions of K-D trees.

Formally, we define a \textbf{K-D tree} built on $n=2^d$ 3D input points $P$ as a full binary tree with $d+1$ \textbf{layers} $L_0,\cdots,L_d$ in a top-down order. There are a total of $2^i$ nodes in $L_i$. A \textbf{leaf node} $o\in L_d$ is corresponding to a unique input $p(o)=p_i$. A \textbf{non-leaf} node $o\in L_i$ where $0\le i<d$ has two exchangeable \textbf{children nodes} $o_l,o_r \in L_{i+1}$, and both of them have a unique \textbf{parent node} $\mathrm{par}(o_l) = \mathrm{par}(o_r)=o$. The structure of a K-D tree can be described with a triple $T=(\{L_i\}, \{(o,\{o_l,o_r\})\}, p(\cdot))$.

For each node $o$, we define its \textbf{sub-tree} $\mathrm{sub}(o)$ as the set containing itself and all nodes in $\mathrm{sub}(o_l)\cup \mathrm{sub}(o_r)$ if $o$ is non-leaf. By induction, we know that if $o\in L_i$, then $|\mathrm{sub}(o)|=2^{d-i}$.

We define the linear \textbf{criterion} $D_o(p)=1_{W_o\cdot p+b_op>0}$ for node $o$, where $W_o$ is a $1\times 3$ matrix and $b_o$ is a scalar. The criterion $D_o$ acts as the criterion in a decision tree's node, and holds $\forall o'\in L_d\cup \mathrm{sub}(o_l), D_o(p(o'))=0$, and $\forall o'\in L_d\cup \mathrm{sub}(o_r), D_o(p(o'))=1$. We call the plane $\{W_op+b_o=0\mid p\}$ the \textbf{division plane} between the left and right children. In the original definition of K-D trees, we limit the division plane to be parallel to an axis plane, or $W_o \in \{(0,0,1), (0,1,0), (1,0,0)\}$.

A node $o$ is corresponding to a continuous 3D \textbf{sub-space} $S(o)$ being the intersection of the criterions of all its ancestor nodes. In the original definition of K-D trees,  all $S(o)$'s are 3D rectangulars. $S(o)$ characterizes the sub-tree of node $o$, so that $S(o)\cap P = \{p(o')\mid o'\in \mathrm{sub}(o) \cup L_d\}$.

As a K-D tree of $2^d$ points is always a full binary tree, it can be determined by the mapping or arrangement $p(o)$ from leaf nodes to input points. The arrangement algorithm $\textbf{arrange-points}(P')\to o$ is a recursive algorithm, which takes a subset $P'$ of input points with $|P'|=2^{d'}$, and returns a depth-$d'$ sub-tree with root node $o$, so that $\mathrm{sub}(o)=P'$. Intuitively, the arrangement algorithm recursively constructs each node of the K-D tree with given input points.

Our algorithm (Algorithm \ref{alg:build-kdt}) acts as below. For a leaf node, the algorithm immediately returns with a single leaf node $o$ with $p(o)=p_i$ s.t. $P'=\{p_i\}$. Otherwise, the algorithm chooses the normal vector $W_o=\textbf{choose-division-plane}(P')$ of the division plane. Then, it finds proper bias $b_o$ to divide $P'$ into two equal-size parts $P'_l$ and $P'_r$, calls $\textbf{arrange-points}$ recursively on each of them, and uses the returned nodes as $o$'s children nodes $o_l$ and $o_r$.

The method $\textbf{choose-division-plane}(P')$ is the key of the whole arrangement algorithm. An implementation of the original K-D tree defines such a method to choose the best axis as the normal vector, according to some metric. In \themodel, we use the principle component of $P'$ (obtained by a PCA algorithm) as the normal vector.

\begin{algorithm}[tb]
   \caption{Build a K-D tree}
   \label{alg:build-kdt}
\begin{algorithmic}[1]
    \Function{Build}{Point cloud $P$}
        \If{$|P|=1$}
            \State \textbf{return} \textbf{make-leaf}($P$)
        \EndIf
        \State $o\gets \textbf{new-node()}$
        \State $W_o \gets \textbf{choose-division-plane()}$
        \State $b_o \gets \text{medium of~} \{W_o\cdot p\mid p\in P \}$
        \State $o_l\gets \text{Build}(\{p\mid W_o\cdot p < b_0, p\in P\})$
        \State $o_r\gets \text{Build}(\{p\mid W_o\cdot p \ge b_0, p\in P\})$
        \State \Return K-D tree rooted at $o$
    \EndFunction
\end{algorithmic}
\end{algorithm}

\section{Transformation Sampling and Dataset Generation}

This section covers more details about the transformed dataset construction mentioned in \textbf{Section 4.1 Transformations} in experiments of the main paper.

\begin{algorithm}[tb]
   \caption{Construct transformed dataset}
   \label{alg:trans-dataset}
\begin{algorithmic}[1]
    \Function{TransData}{Dataset $D$, Transformation distribution $T$, Augment time $a$}
        \State $D_t\gets \emptyset$
        \ForAll{$P\in D$}
            \Repeat
                \State Sample transformation $t\sim T$
                \State $D_t \gets D_t\cap \{t(P)\}$
            \Until{repeated $a$ times}
        \EndFor
        \State \Return $D_t$
    \EndFunction
\end{algorithmic}
\end{algorithm}

We use Algorithm \ref{alg:trans-dataset} to generate our transformed dataset from an existing dataset $D$ with transformation distribution $T$. In this algorithm, each single data i.e., point cloud in $D$ will be applied with $a$ (augment time) {\em different} transformations in $T$ and included in the transformed dataset.

We define 3 transformation distributions: $T_{\mathrm{affine}}$,  $T_{\mathrm{affine\_agg}}$, and $T_{\mathrm{projective}}$. $T_{\mathrm{affine}}$ generates a random affine transformation with \texttt{torch.nn.Linear(3, 3, bias=False)} in PyTorch, i.e., generating a random $3\times 3$ affine matrix $A$ where $A_{i,j}\sim U\left(-\frac{1}{\sqrt{3}},\frac{1}{\sqrt{3}}\right)$. $T_{\mathrm{affine\_agg}}$ is the distribution of the affine transformation with maximum EAD (a metric for measuring transformation intensity, defined in \textbf{Section 3.2 Robustness Against Transformations} in the main paper) among 5,000 different samples from $T_{\mathrm{affine}}$, which represents more aggressive affine transformations with larger EAD. 

$T_{\mathrm{projective}}$ is defined with a special generating algorithm to guarantee that there is no numerical issue. The point cloud is unitilized to $[-1,1]^3$, and then applied with a random affine transformation sampled from $T_{\mathrm{affine}}$. Then, we randomly select 4 points from $[-2,2]^3$, and use three of them as vanishing points $V_x,V_y,V_z$ for each axis and the remaining one as the point $O_p$ which the original point $O$ is projected to. Then, we randomly generate scalor arguments $a,b,c$ and decide $d$ to construct the projective matrix as below \begin{equation} \begin{pmatrix}
aV_x & a\\
bV_y & b \\
cV_z & c \\
dO_p & d
\end{pmatrix}. \end{equation} For the argument $d$, we randomly select it in the range that makes sure that no points will be projected to a point with infinity coordinates.

\section{Robustness Against Similarity Transformations}

In \textbf{Section 3.2 Robustness Against Transformations} in the main paper, we analyzed the robustness against similarity transformations. We  explain more details and prove the lemma in this section.

A \textbf{similarity transformation} is a transformation that preserves the shape (degree of all angles) of a point cloud. It includes rotations, flips, scales, and shifts.

\themodel uses relaxed K-D trees as the base tree, which holds lemma below:

\textbf{Lemma.} If $\textbf{choose-division-plane}(P)$ is equivariant to a similarity transformation $\sigma$, or \begin{equation}\textbf{choose-division-plane}(\sigma(P)) = \sigma(\textbf{choose-division-plane}(P)),\end{equation} then the whole arrangement algorithm $\textbf{arrange-points}(P)$ is equivariant to such similarity transformation. Also, each leaf's corresponding point $p(o)$ is invariant to similarity transformation $\sigma$.

\textbf{Proof.}

The normal vectors of division planes returned by $\textbf{choose-division-plane}(P)$ is agnostic to shift and global scales, since both transformations do not change the direction of such planes indicated by normal vectors. As a result, $\textbf{choose-division-plane}(P)$ is natively invariant to shift and scaling. Equivalently, we can assume the input point cloud $P$ 
is already re-centered (shift to make the center of mass locates at $O$) and unitilized (scale to make all points locate in a unit sphere and one point is on the sphere). In this way, we do not need to consider the shift and scaling components in $\sigma$, and $\sigma$ can therefore be regarded as a orthogonal $3\times 3$ affine matrix. 

If the division plane $W_o$ returned by $\textbf{choose-division-plane}(P)$ is equivariant to the transformation $\sigma$, the division plane in the same function call in the K-D tree construction of $\sigma(P)$ will choose the division plane $\sigma(W_o)$. For each point $p\in P$, the criterion for division into $P^o_l=\{p\mid W_o\cdot p < b_0, p\in P\}$ or $P^o_r = \{p\mid W_o\cdot p \ge b_0, p\in P\}$ is the value of $W_o\cdot p$; and for $\sigma(p)\in\sigma(P)$, the criterion is (here we regard $p$ and $W_o$ as column vectors) \begin{equation}\sigma(W_o)\cdot\sigma(p) = W_o^\top \sigma^\top\sigma p=W_o^\top p=W_o\cdot p,\end{equation} which is the same as that of the construction of the original point cloud $P$. So, the divided point sets $P^o_l$ and $P^o_r$ will be equivariant to transformation $\sigma$, i.e., if $p$ is divided into $P^o_l$ in construction of $P$, then it will also be divided into $P^o_l$ in construction of $\sigma(P)$, and vice versa for $P^o_r$. They will be passed to the recursive call of \textbf{Build($\cdot$)} at line 8 and 9 in Algorithm \ref{alg:build-kdt}. By induction, each function call of \textbf{Build($\cdot$)} will be exactly the same for $P$ and $\sigma(P)$, so that the division plane chosen by each function call will be equivariant to $\sigma$, and the structure of the K-D tree $T=(\{L_i\}, \{(o,\{o_l,o_r\})\}, p(\cdot))$ will be the invariant.

\themodel uses PCA to implement $\textbf{choose-division-plane}(P)$, which is equivariant under any similarity transformation. As a result, the structure of the relaxed K-D tree $T=(\{L_i\}, \{(o,\{o_l,o_r\})\}, p(\cdot))$ remains identical after applying the similarity transformation $\sigma$, and the bottom-up flow also works in an identical way. This makes our model's working flow invariant to similarity transformations.

\section{Robustness Against Affine Transformation}
Our pre-alignment method and the function \textbf{choose-division-plane}() in K-D tree construction use PCA, which is invariant to similarity transformations, as discussed in the previous section and in \textbf{Section 3.2 Robustness Against Transformations} in the main paper. In this section, we discuss more about the robustness against affine transformations. 

There is some work~\cite{affinvdesc,affinvfunc} that uses PCA to design affine-invariant descriptors or functions for 2D and 3D point sets. It has been theoretically proved that some features obtained by PCA (e.g., based on eigenvalues and eigenvectors) are affine-invariant. Such an observation shows the potential of PCA in extracting affine-invariant/robust information -- we hypothesize that this is the underlying reason that our pre-alignment and \themodel with PCA-based relaxed K-D tree construction are robust against affine transformations. We leave a rigid proof as interesting future work.

We also provide empirical analysis for pre-alignment. We calculate EAD (a metric for measuring transformation intensity, defined in \textbf{Section 3.2 Robustness Against Transformations} in the main paper) on pre-aligned original point clouds and pre-aligned affine-transformed point clouds, and the results are shown in Table \ref{tab:exp-pa-ead}. We can find that the mean EAD is smaller than $10^{-4}$ in all affine cases, which means that the shape after pre-alignment is invariant to affine transformations with high precision. The experiments also show that our pre-alignment can highly reduce the EAD for projective-transformed point clouds. These results empirically shows that our pre-alignment is highly invariant to affine transformations and can also work on projective transformations, and thus it is a simple yet powerful and general approach for pre-processing of point clouds.

\begin{table}[t]
\centering

\caption{Our pre-alignment (`PA') method achieves a mean EAD that is very close to zero in affine transformed point clouds, and highly reduces the EAD of projective transformed point clouds by $50\%$. For each number in the table, we randomly sample $3,000$ transformations and take the mean of the EAD between pre-aligned original and pre-aligned transformed point clouds. The point clouds with a label (e.g., `Laptop') are taken from ModelNet40 train data, and we also provide their index in the training dataset like \#0. For random point clouds, the point sets are uniformly sampled from $[-1,1]^{2048\times 3}$}
\setlength{\tabcolsep}{4pt}
\scalebox{0.85}{\begin{tabular}{l|cccc}
 \hline
 Point Cloud & Affine & Affine Aggressive & Projective & Projective w/o PA \\
 \hline
Laptop (\#0) & $8\times 10^{-7}$ & $5\times 10^{-6}$ &0.3073 &0.9641 \\
Wardrobe (\#1) &$9\times 10^{-7}$ &$2\times 10^{-6}$ &0.3377 &0.9191\\
Table (\#5)  & $1\times 10^{-6}$ & $3\times 10^{-6}$ &0.2979 &0.8996\\
Airplane (\#11) &$7\times 10^{-7}$ &$8\times 10^{-6}$ &0.3254 &0.9806 \\
\hline
Random &$4\times 10^{-7}$ &$2\times 10^{-6}$ &0.6745 &1.2173 \\
 \hline
 
\end{tabular}}
\label{tab:exp-pa-ead}
\vspace{-3pt}
\end{table}

\section{Details About Segmentation Component}
This section covers more details about the general segmentation component in \textbf{Section 3.3 Downstream Components} in the main paper.

First, we introduce the top-down information flow, which is opposite to the bottom-up flow we used in the encoder. Each node has an information $\mathrm{self}(o)$ of itself, and we compute the carried information $\mathrm{carry}(o)$ that is downloaded from its ancestors. 

The input of such an information flow is $\{\mathrm{self}(o)\}$ and $\mathrm{carry}(R)=\mathrm{self}(R)$ for the root node $R$. Then, for each node $o$, its $\mathrm{carry}(o)$ is obtained by aggregating $\mathrm{self}(o)$ and its parent's carried information $\mathrm{carry}(\mathrm{par}(o))$, as the formula below:
\begin{equation}
\mathrm{carry}(o) = \textbf{merge-carry}(\mathrm{self}(o), \mathrm{carry}(\mathrm{par}(o))).
\end{equation}
The carried information contains all information of its ancestor node. For example, if we want to compute the number of ancestors for each node, we can set $\mathrm{self}(o)=1$ for all nodes, and 
\begin{equation}
\textbf{merge-carry}(a,b) = a+b.
\end{equation}
After running the top-down information flow, $\mathrm{carry}(o)$ will be the number of ancestors (including itself) for each node $o$.

\begin{figure}[t]

\begin{center}
\centerline{\includegraphics[width=0.7\columnwidth]{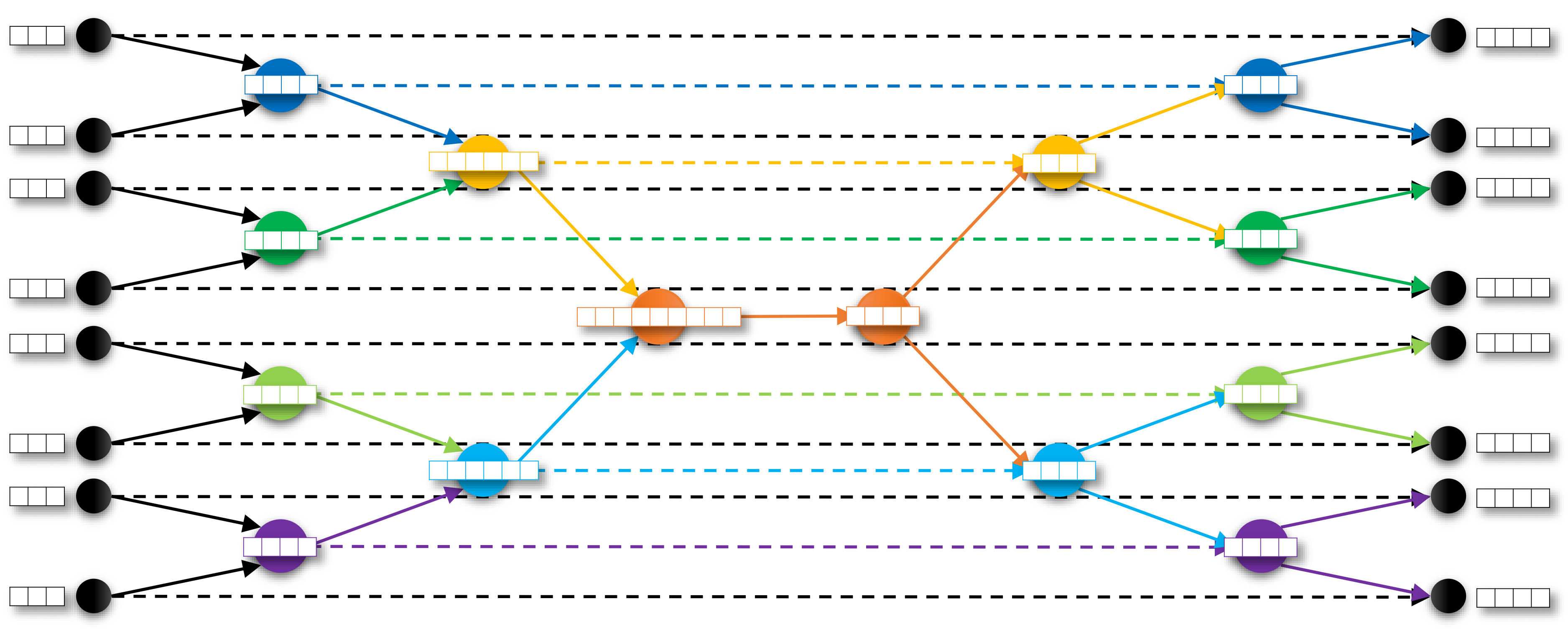}}
\caption{The segmentation model. The two symmetric K-D trees represent two stages: a bottom-up information flow (\textbf{left}) and a top-down information flow (\textbf{right}), on the same K-D tree}
\label{fig:segment-2}
\end{center}
\end{figure}

For the segmentation task, we maintain both a bottom-up flow and a top-down flow on the same K-D tree (as shown in Figure \ref{fig:segment-2}) and define $\mathrm{self}(o) = \mathrm{info}(o)$ with a skip connection from the symmetric node in the bottom-up information flow. As $\mathrm{carry}(R) = \mathrm{self}(R) = \mathrm{info}(R)$ is the vector of global features of the point cloud, intuitively, in each $\textbf{merge-carry}$ step, we are obtaining a connection between the local information and the global information. For each leaf node $o$, $\mathrm{carry}(o)$ can be regarded as a connection between the point $p(o)$ and the whole point cloud . So we can regard such $\mathrm{carry}(o)$ as the point feature of point $p(o)$. We use an MLP to classify each leaf node's point feature, and output the log-likelihood score for each segment class candidate.

\section{Part Segmentation on Projective Transformed Dataset}

The segmentation results on projective transformed ShapeNetPart~\cite{shapenet} is in Table \ref{tab:exp-seg-proj}.  Our \themodel achieves a top mIoU over all the baselines with an increment of nearly 10\%, and achieves the state-of-the-art mean IoU in 75\% of all classes. It shows that \themodel outperforms other baselines with a larger gap under more challenging projective transformation.

\begin{table}[t]
\centering
\caption{On projectived transformed, pre-aligned ShapeNetPart~\cite{shapenet}, \themodel significantly outperforms the baselines in class-level mIoU(\%), and achieves the state-of-the-art mean IoU in 75\% of all classes}

\setlength{\tabcolsep}{1pt}
\scalebox{0.73}{\begin{tabular}{l|cccccccccccccccc|c}
 \hline
 Method & {\tiny airplane} & {\tiny bag} & {\tiny cap} & {\tiny car} & {\tiny chair} & {\tiny earphone} & {\tiny guitar} & {\tiny knife} & {\tiny lamp} & {\tiny laptop} & {\tiny motorbike} & {\tiny mug} & {\tiny pistol} & {\tiny rocket} & {\tiny skateboard} & {\tiny table} &  mIoU \\
 
 \hline 
 DGCNN~\cite{dgcnn} & \bf{69.8}& 46.5& \bf{92.4}& 29.3& 36.8& \bf{92.0}& 83.7& \bf{75.6}& 65.9& \bf{80.6}& 14.5& 45.7& 10.4& 39.1& 29.8& 66.8&  55.0  \\
 CurveNet~\cite{curvenet} &22.4 &44.8 &36.6 &5.7 &30.7 &20.5 &22.7 &25.2 &45.9 &27.2 &16.9 &48.6 &23.7 &22.7 &43.7 &45.5 & 30.2 \\
 \hline
 \themodel Def &67.4 &\bf{63.0} &76.4 &\bf{49.0} &\bf{75.2} &57.1 &\bf{84.2} &70.7 &\bf{67.7} &59.9 &\bf{39.9} &\bf{80.8} &\bf{60.3} &\bf{52.5} &\bf{57.8} &\bf{72.2} &\bf{64.6}  \\
 \hline
\end{tabular}}
\label{tab:exp-seg-proj}
\vspace{-3pt}
\end{table}

\section{Architectures of \themodel Variants}
\begin{figure}[t]
\begin{center}
\textbf{(a) \themodel Def}
\centerline{\includegraphics[width=0.9\columnwidth]{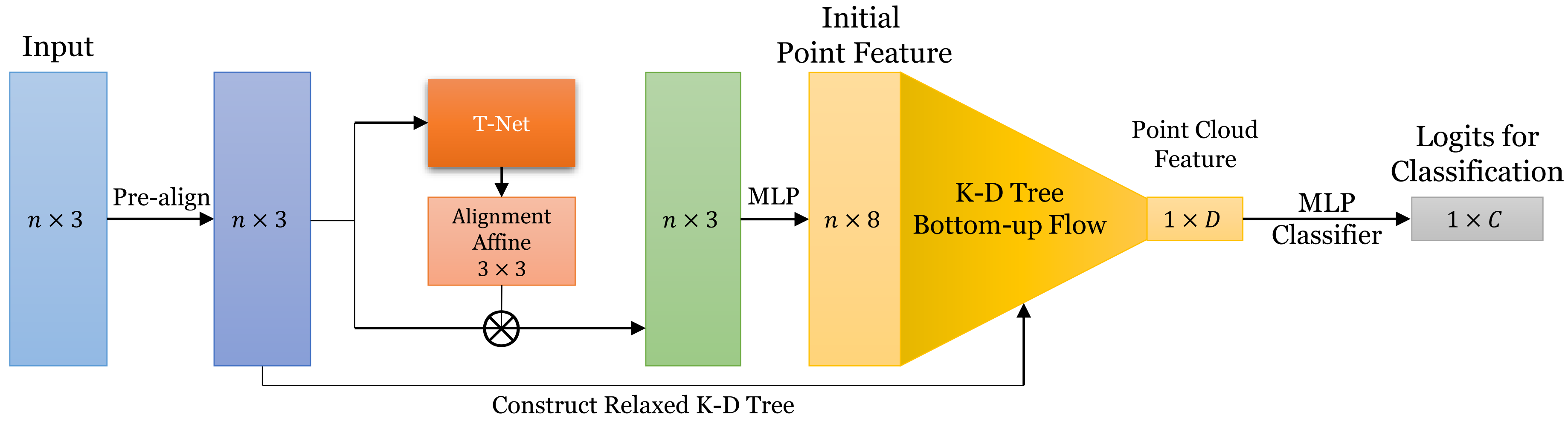}}
\textbf{(b) \themodel KA}
\centerline{\includegraphics[width=0.9\columnwidth]{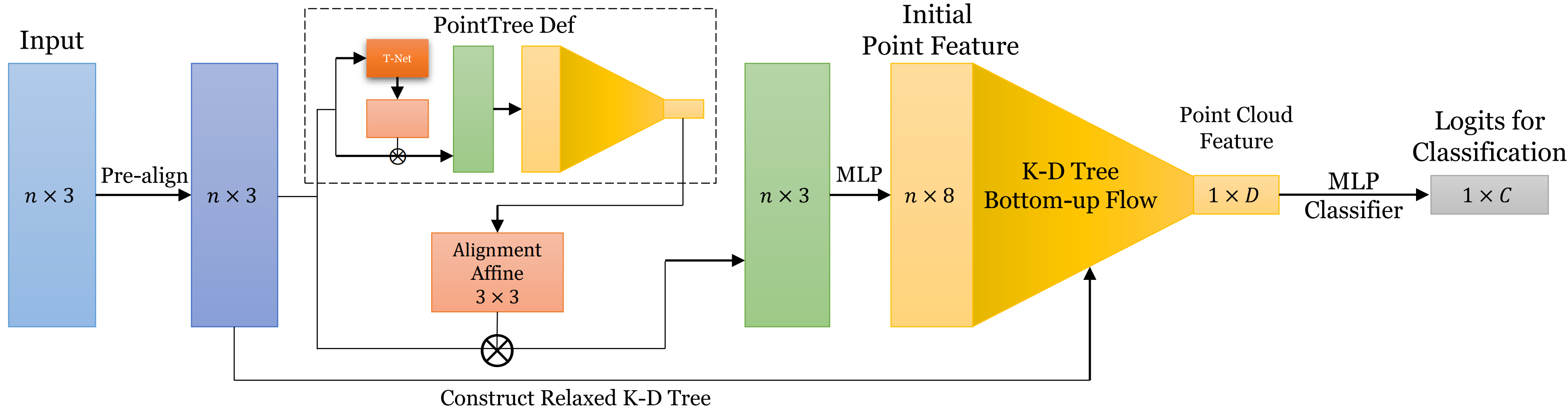}}
\textbf{(c) \themodel RNS}
\centerline{\includegraphics[width=0.9\columnwidth]{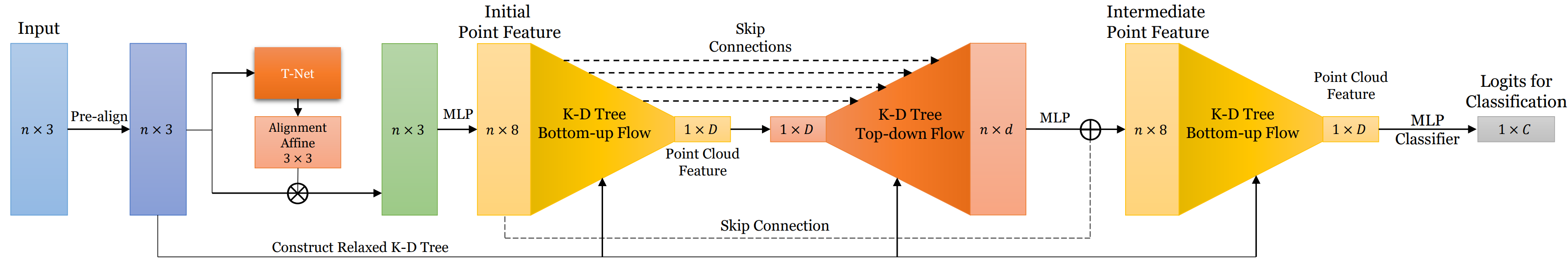}}
\caption{The architectures of 3 \themodel Variants: (a) Def, (b) KA, and (c) RNS. (a) \themodel Def is the simplest model that uses the T-Net in PointNet~\cite{pointnet} as the alignment network. (b) \themodel KA replaces the T-Net with another \themodel Def, and uses its output, the point cloud feature, to generate the $3\times 3$ alignment affine matrix. (c) \themodel RNS connects a \themodel Def with a segmentation component, and uses the segmentation component's output plus the skip connection from the initial point feature as the input features for another K-D tree bottom-up flow}
\label{fig:arch-var}
\end{center}
\end{figure}

According to \textbf{Section 4.2 Baselines and PointTree Variants} in the main paper, we have 3 variants of \themodel: Def, KA, and RNS. Their architectures are shown in Figure \ref{fig:arch-var}.

For \themodel KA, we design this variant by introducing a stronger alignment network based on \themodel Def encoder. As we mentioned in ``Alignment Network'' in Section~\ref{sec:robustsub}, the alignment network supports any encoder that outputs a point cloud feature. Different from the default encoder that uses T-Net in PointNet~\cite{pointnet} as the alignment network, we build the alignment network as another \themodel Def encoder in this version.

For \themodel RNS, we can stack multiple ResNet blocks to get a larger model. We only use the single ResNet block version in all our experiments.

\section{Additional Implementation Details}

In this section, we include additional implementation details when conducting experiments in \textbf{Section 4 Experiments} in the main paper.

\textbf{Hyperparameters.} The architectures of the model variants we use are shown in Figure~\ref{fig:arch-var}. We use 2,048 points in each point cloud, and build a 12-level tree. In the bottom-up flow, the dimension of node features from bottom to top is 32, 64, 128, 256, 512, 512, 1024, 1024, 2048, 2048, 4096, and 4096. In the top-down flow, the carried information's dimension is 512. The batch size is set to 32 for all segmentation tasks and for \themodel RNS on classification tasks, and 64 for all the other experiments.

\textbf{Implementation.} We implement the whole \themodel with PyTorch. The PCA in pre-alignment and K-D tree construction are implemented with a library function \texttt{torch.pca\_lowrank}. For the dimension-increasing MLP at each layer, we use one \texttt{torch.nn.Linear}. For the MLP classifier for both classification and segmentation, we use a 3-layer MLP with ReLU activation and batch normalization for hidden layers.

\textbf{Data Augmentation.} When training \themodel, we do data augmentation by applying random axis flipping and axis permutations. And, for pre-aligned experiments, we also augment the data by applying random affine transformations on the point cloud then pre-align again. According to Figure \ref{fig:arch-var}, these augmentations are inserted after pre-alignment and only affect the coordinate inputs of \themodel. They do {\em not} affect the construction of K-D trees. The K-D tree for each point cloud is constructed only on the transformed coordinates (with or without pre-alignment in different settings).

\textbf{Training.} We run all experiments of \themodel with a PC on NVIDIA RTX 3060 GPU. We train the model with mini-batch training and use Adam as the optimizer. It took 20 hours for full convergence, but according to ``Overtime Accuracy'' in \textbf{Section 4.3 Classification on ModelNet40} in the main paper, it can achieve a comparable result within 2.5 hours. For ModelNet40~\cite{modelnet}, it only includes train and test splits. We further split the original training dataset into the new training dataset and the validation dataset. We train \themodel only on the new train dataset and tune hyperparameters only on the validation dataset. For ShapeNetPart~\cite{shapenet}, it already have the splits of training, validation, and test, so we train \themodel using these splits in the common way.

\section{Iterative Pre-alignment}

There is still some issues in our PCA-based pre-alignment. The solution of PCA is not unique (e.g., for a sphere-shaped point cloud, any vector is the principal component). Though our experiments shows that the EAD between differently affine transformed point cloud is very small, i.e., the they have same shapes, there may still be a similarity transformation between them. Also, the scaling factor $\Sigma$ provided by PCA may be not optimal for our model to obtain best results.

To obtain better results, we further propose an iterative pre-alignment scheme, to iterative align the point cloud with PCA and apply a designed scaling factor until convergence. The algorithm is shown in Algorithm \ref{alg:iter-pa}.

\begin{algorithm}[tb]
   \caption{Iterative Pre-alignment}
   \label{alg:iter-pa}
\begin{algorithmic}[1]
    \Function{IterativePreAlign}{Point cloud $P$, Maximum Iteration $m$}
        \While{\# Iterations $<m$}
            \State $U,\Sigma, V\gets \textbf{PCA}(P)$
            \If{Vectors in $\{V\}$ are x,y,z-axes}
                \State (\textit{The axes suggested by PCA converges.})
                \State \textbf{break}
            \EndIf
            \State (\textit{Align the point cloud with PCA.})
            \State $P\gets U$
            \State (\textit{Apply a designed scaling scheme.})
            \For{$\textbf{d}\in {\textbf{x},\textbf{y},\textbf{z} }$}
                \State (\textit{For each axis, unitilize the coordinates to make the average abs. 1.})
                \State $f\gets \textbf{Mean}_{p\in P} |\textbf{d}_p|$
                \State $\textbf{d}_p\gets \textbf{d}_p / f$, $\forall p\in P$
            \EndFor
        \EndWhile
        \State \Return $P$
    \EndFunction
\end{algorithmic}
\end{algorithm}

The iterative pre-alignment can stabilize and improve the performance of \themodel.

\section{Overtime Accuracy of Classification Task}
\begin{figure}[t]
\begin{center}
\centerline{\includegraphics[width=0.9\columnwidth]{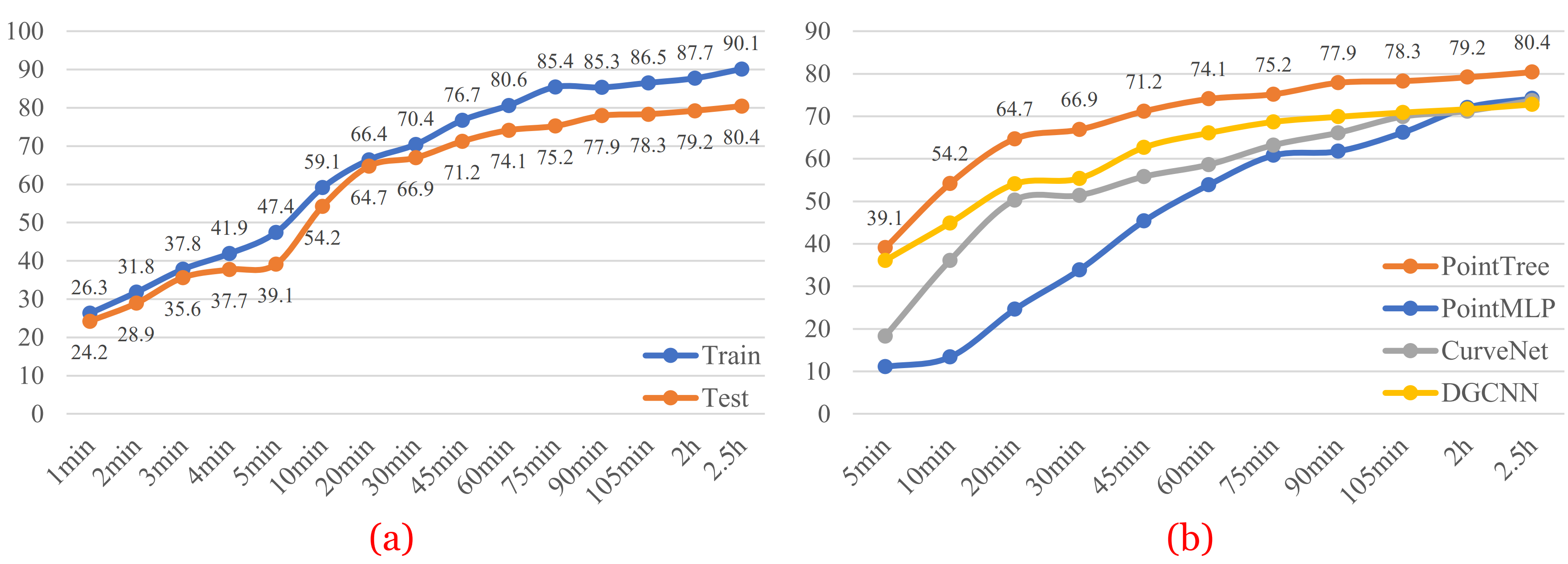}}
\caption{\themodel converges very fast during training time and leads the test accuracy among all baselines. It highly adapts the transformed data so that it can achieve a reasonable accuracy within 5 minutes, and reaches 95 percent of its best accuracy (\%) in a very limited time of 2.5 hours. (a) The overtime train and test accuracy (\%) of our most powerful model, \themodel RNS. (b) The overtime test accuracy of \themodel and top baselines}
\label{fig:exp-time-accu}
\end{center}
\end{figure}

Figure \ref{fig:exp-time-accu}\textcolor{red}{-a} shows the train and test accuracy of our most powerful model, \themodel RNS, on pre-aligned ModelNet40 with affine transformation. Even with very limited time of 2.5 hours, our model can still achieve 95 percent of its best accuracy (80.4\% out of 84.1\%), which already outperforms many baseline models. \themodel can even achieve a reasonable accuracy of 24.2\% within 1 minute. These facts show that \themodel is highly adaptive on transformed data. Figure \ref{fig:exp-time-accu}\textcolor{red}{-b} shows  overtime test accuracy of \themodel and top baselines. Among these models, \themodel is always leading the test accuracy, and has a gap that is more than 5\% higher than all these baselines at most of the time.

\section{Stability Test of \themodel}

\begin{table}[t]
\centering
\caption{\themodel has stable performance with a low standard variance over different random transformations on each transformed ModelNet40. Its accuracy on all experiments is higher than the baseline PointMLP~\cite{pointmlp} with statistical significance. The second column is the mean EAD over all point clouds in the dataset (a more difficult task has a higher EAD). The accuracy is also instance-level accuracy (\%), and the standard variance $\sigma$(\%) are computed over ModelNet40's testing data with different random transformations}
\setlength{\tabcolsep}{8pt}
\scalebox{0.85}{\begin{tabular}{l|cc|c}
 \hline
 Transformation &  Mean EAD & PointMLP~\cite{pointmlp} & \themodel RNS\\
 \hline
    Affine w/ PA & $10^{-4}$ & 82.3 & $84.1^{\sigma=0.23}$  \\
    Affine w/o PA & $0.387$ & 63.1 & $73.1^{\sigma=0.22}$ \\
    Affine (Aggressive) w/o PA & $0.785$ & 37.5 &  $63.7^{\sigma=0.38}$ \\
\hline
    Projective w/ PA &$0.307$ & 49.9 & $62.1^{\sigma=0.67}$\\
    Projective w/o PA & $0.960$ & 4.1 & $31.8^{\sigma=0.76}$\\
 \hline
 
\end{tabular}}
\label{tab:exp-cls-ours}
\end{table}

 Table \ref{tab:exp-cls-ours} shows the stability test results. In this experiment, we sample multiple transformations for each point cloud in test data, and compute the standard variance to show the stability of our model on different transformations. Also, we add a special test: ``affine (aggressive)'', for which we sample affine transformation from another distribution that is expected to have higher EAD. The experiment results show that (1) \themodel is stable among different transformations with low standard variance and (2) \themodel's accuracy is higher than the baseline PointMLP~\cite{pointmlp} with statistical significance.

\section{Details About S3DIS}

 We evaluate \themodel for the point cloud semantic segmentation on S3DIS~\cite{s3dis}. It contains 6 areas with 271 rooms. There is a point cloud for each room, containing all objects' surfaces in the room within 13 object categories, e.g., ceiling, floor, window, etc. In our evaluation, we use the PointNet~\cite{pointnet} version of S3DIS, which ``sample(s) rooms into blocks with area 1m by 1m'', and ``randomly sample(s) 4,096 points in each block''. We use the processed data released on PointNet's Github repository for training and evaluation, to make it a fair comparison over all baselines.

\bibliographystyle{splncs04}
\bibliography{main}

\end{document}